\documentclass[10pt,twocolumn,letterpaper]{article}

\usepackage{iccv}
\usepackage{times}
\usepackage{epsfig}
\usepackage{graphicx}
\usepackage{amsmath}
\usepackage{amssymb}
\usepackage{soul}
\usepackage{array,multirow}
\usepackage{algorithm}
\usepackage{algorithmic}

\usepackage[lofdepth,lotdepth]{subfig}
\usepackage{float}
\usepackage{paralist}
\usepackage{caption}
\usepackage{multirow}
\usepackage{authblk}

\usepackage[pagebackref=true,breaklinks=true,letterpaper=true,colorlinks,bookmarks=false]{hyperref}

\newcommand{\cQ}{\mathcal{Q}}
\newcommand{\cG}{\mathcal{G}}

\newcommand{\cC}{\mathcal{C}}
\newcommand{\cD}{\mathcal{D}}

\newcommand{\bE}{\mathbf{E}}
\newcommand{\cE}{\mathcal{E}}

\newcommand{\bL}{\mathbf{L}}

\newcommand{\bO}{\mathbf{O}}

\newcommand{\bp}{\mathbf{p}}

\newcommand{\bX}{\mathbf{X}}

\newcommand{\cN}{\mathcal{N}}
\newcommand{\bK}{\mathbf{K}}

\newcommand{\cM}{\mathcal{M}}

\newcommand{\cV}{\mathcal{V}}

\DeclareMathOperator*{\argmax}{arg\,max}

\iccvfinalcopy 

\def\iccvPaperID{3667} 

\ificcvfinal\fi
\begin{document}

\title{Scalable Place Recognition Under Appearance Change for Autonomous Driving}

\author[1]{Anh-Dzung Doan}
\author[1]{Yasir Latif}
\author[1]{Tat-Jun Chin}
\author[1]{Yu Liu}
\author[2]{Thanh-Toan Do}
\author[1]{Ian Reid}
\affil[1]{School of Computer Science, The University of Adelaide}
\affil[2]{Department of Computer Science, University of Liverpool}
\maketitle

\begin{abstract}
 A major challenge in place recognition for autonomous driving is to be robust against appearance changes due to short-term (e.g., weather, lighting) and long-term (seasons, vegetation growth, etc.) environmental variations. A promising solution is to continuously accumulate images to maintain an adequate sample of the conditions and incorporate new changes into the place recognition decision. However, this demands a place recognition technique that is scalable on an ever growing dataset. To this end, we propose a novel place recognition technique that can be efficiently retrained and compressed, such that the recognition of new queries can exploit all available data (including recent changes) without suffering from visible growth in computational cost. Underpinning our method is a novel temporal image matching technique based on Hidden Markov Models. Our experiments show that, compared to state-of-the-art techniques, our method has much greater potential for large-scale place recognition for autonomous driving.
\end{abstract}

\vspace{-0.8em}
\section{Introduction}
\vspace{-0.5em}

Place recognition (PR) is the broad problem of recognizing ``places" based on visual inputs \cite{lowry2016survey, visuallocaizationtutorialECCV2018}. Recently, it has been pursued actively in autonomous driving research, where PR forms a core component in localization (i.e., estimating the vehicle pose)~\cite{sattler2017efficient,kendall2015posenet,brachmann2017dsac,clark2017vidloc,sattler2018benchmarking,brachmann2018learning,brahmbhatt2018mapnet} and loop closure detection~\cite{FABMAP,BBOW}. Many existing methods for PR require to train on a large dataset of sample images, often with ground truth positioning labels, and state-of-the-art results are reported by methods that employ learning~\cite{kendall2015posenet,kendall2017geometric,brahmbhatt2018mapnet,clark2017vidloc}.


To perform convincingly, a practical PR algorithm must be robust against appearance changes in the operating environment. These can occur due to higher frequency environmental variability such as weather, time of day, and pedestrian density, as well as longer term changes such as seasons and vegetation growth. A realistic PR system must also contend with ``less cyclical" changes, such as construction and roadworks, updating of signage, fa\c{c}ades and billboards, as well as abrupt changes to traffic rules that affect traffic flow (this can have a huge impact on PR if the database contains images seen from only one particular flow \cite{FABMAP,BBOW}). Such appearance changes invariably occur in real life.

To meet the challenges posed by appearance variations, one paradigm is to develop PR algorithms that are inherently robust against the changes. Methods under this paradigm attempt to extract the ``visual essence" of a place that is independent of appearance changes \cite{arandjelovic2016netvlad}. However, such methods have mostly been demonstrated on more ``natural" variations such as time of day and seasons.

Another paradigm is to equip the PR algorithm with a large image dataset that was acquired under different environmental conditions \cite{churchill2013experience}. To accommodate long-term evolution in appearance, however, it is vital to continuously accumulate data and update the PR algorithm. To achieve continuous data collection cost-effectively over a large region, one could opportunistically acquire data using a fleet of service vehicles (e.g., taxis, delivery vehicles) and amateur mappers. Indeed, there are street imagery datasets that grow continuously through crowdsourced videos~\cite{neuhold2017mapillary,haklay2008openstreetmap}. Under this approach, it is reasonable to assume that a decent sampling of the appearance variations, including the recent changes, is captured in the ever growing dataset.

Under continuous dataset growth, the key to consistently accurate PR is to ``assimilate" new data quickly. This demands a PR algorithm that is \emph{scalable}. Specifically, the computational cost of testing (i.e., performing PR on a query input) should not increase visibly with the increase in dataset size. Equally crucially, updating or retraining the PR algorithm on new data must also be highly efficient.

Arguably, PR algorithms based on deep learning \cite{brahmbhatt2018mapnet,clark2017vidloc} can accommodate new data by simply appending it to the dataset and fine-tuning the network parameters. However, as we will show later, this fine-tuning process is still too costly to be practical, and the lack of accurate labels in the testing sequence can be a major obstacle.

\vspace{-1em}
\paragraph{Contributions}

We propose a novel framework for PR on large-scale datasets that continuously grow due to the incorporation of new sequences in the dataset. To ensure scalability, we develop a novel PR technique based on Hidden Markov Models (HMMs) that is lightweight in both training and testing. Importantly, our method includes a topologically sensitive compression procedure that can update the system efficiently, \emph{without} using GNSS positioning information or computing visual odometry. This leads to PR that can not only improve accuracy by continuous adaption to new data, but also maintain computational efficiency. We demonstrate our technique on datasets harvested from Mapillary~\cite{neuhold2017mapillary}, and also show that it compares favorably against recent PR algorithms on benchmark datasets.

\section{Problem setting}\label{sec:setting}

We first describe our adopted setting for PR for autonomous driving. Let $\mathcal{D} = \begin{Bmatrix}\mathcal{V}_1, \dots, \mathcal{V}_M \end{Bmatrix}$ be a dataset of $M$ videos, where each video
\begin{align}\label{eq:training}
\mathcal{V}_i = \left\{ I_{i,1}, I_{i,2}, \dots, I_{i,N_i} \right\} = \{ I_{i,j} \}^{N_i}_{j=1}
\end{align}
is a time-ordered sequence of $N_i$ images. In the proposed PR system, $\mathcal{D}$ is collected in a distributed manner using a fleet of vehicles instrumented with cameras. Since the vehicles could be from amateur mappers, accurately calibrated/synchronized GNSS positioning may not be available. However, we do assume that the camera on all the vehicles face a similar direction, e.g., front facing. The query video is represented as
\begin{align}
\mathcal{Q} = \left\{ Q_1, Q_2, \dots, Q_T \right\}
\end{align}
which is a temporally-ordered sequence of $T$ query images. The query video could be a new recording from one of the contributing vehicles (recall that our database $\mathcal{D}$ is continuously expanded), or it could be the input from a ``user" of the PR system, e.g., an autonomous vehicle.

\vspace{-0.6em}
\paragraph{Overall aims}

For each $Q_t \in \mathcal{Q}$, the goal of PR is to retrieve an image from $\mathcal{D}$ that was taken from a similar location to $Q_t$, i.e., the FOV of the retrieved image overlaps to a large degree with $Q_t$. As mentioned above, what makes this challenging is the possible variations in image appearance.

In the envisioned PR system, when we have finished processing $\mathcal{Q}$, it is appended to the dataset
\begin{align}
\mathcal{D} = \mathcal{D} \cup \{ \mathcal{Q} \},
\end{align}
thus the image database could grow unboundedly. This imposes great pressure on the PR algorithm to efficiently ``internalise" new data and compress the dataset. As an indication of size, a video can have up to 35,000 images.

\begin{figure*}
	
	\centering
	
	\mbox
	{
		\subfloat[][]
		{
			\includegraphics[width=0.2\textwidth]{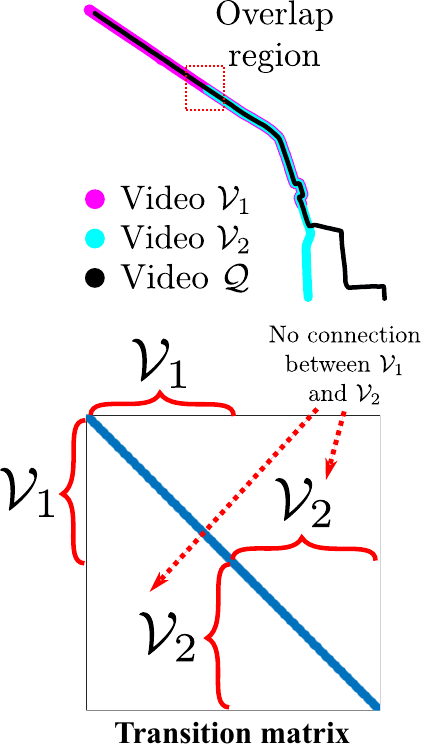}
			
			\label{fig:HMM_new_connection_1}
		}
		
		\hspace{0.15cm}
		
		\subfloat[][]
		{
			
			\includegraphics[width=9.5cm,height=7cm]{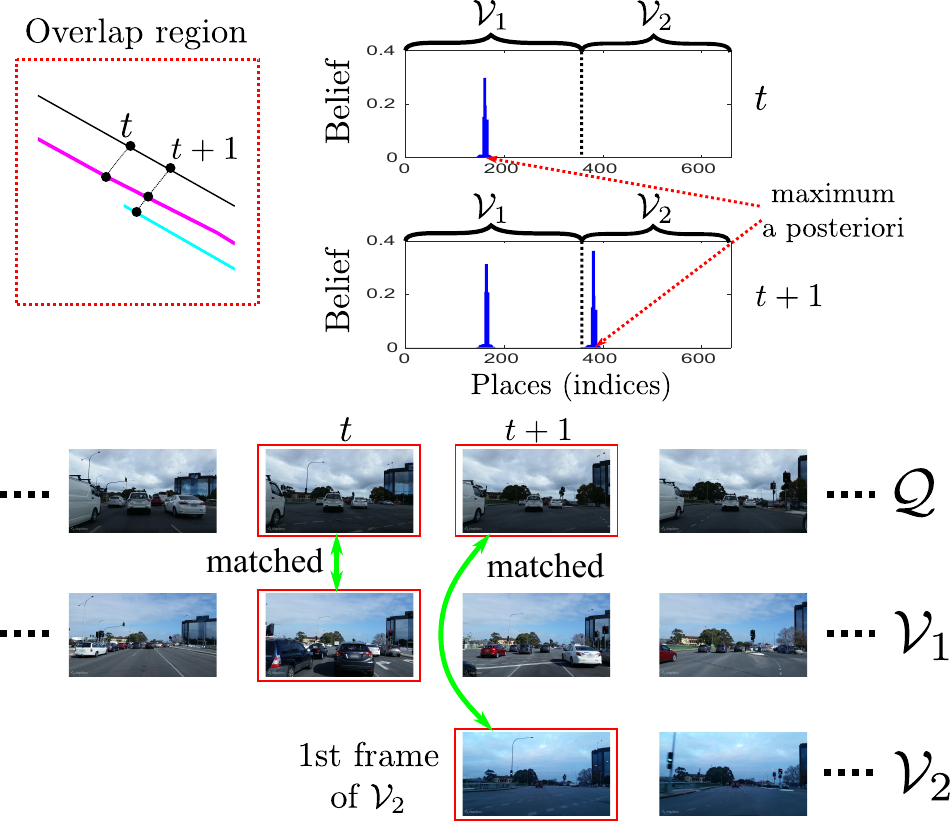}
			
			\label{fig:HMM_demo}
		}
		
		\hspace{0.15cm}
		
		\subfloat[][]
		{
			\includegraphics[width=0.2\textwidth]{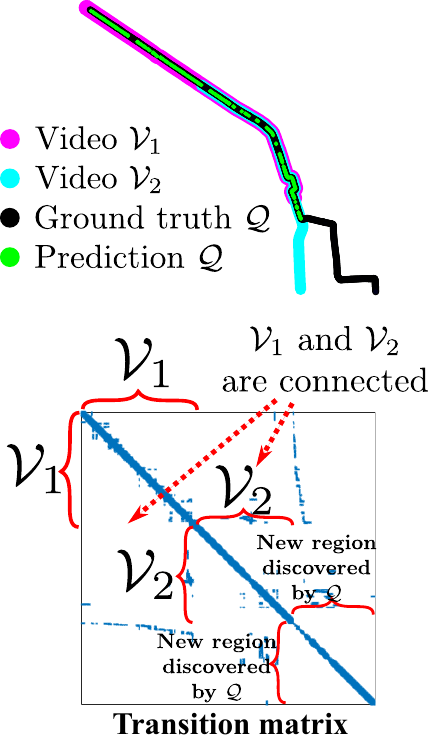}
			
			\label{fig:HMM_new_connection_2}
		}
		
	}
	
	\caption{An overview of our idea using HMM for place recognition. Consider dataset $\cD = \{ \cV_1, \cV_2 \}$ and query $\cQ$. Figure \ref{fig:HMM_new_connection_1}: Because $\cV_1$ and $\cV_2$ are recorded in different environmental conditions, $\cV_2$ cannot be matched against  $\cV_1$, thus there is no connection between $\cV_1$ and $\cV_2$. Query $\mathcal{Q}$ visits the place covered by $\cV_1$ and $\cV_2$, and then an unknown place. Figure \ref{fig:HMM_demo}: Query $\cQ$ is firstly localized against only $\cV_1$. When it comes to the ``Overlap region" at time $t+1$, it localizes against both $\cV_1$ and $\cV_2$. The image corresponding to MaxAP at every time step $t$ is returned as the matching result. Figure \ref{fig:HMM_new_connection_2}: A threshold decides if the matching result should be accepted, thus when $\c{Q}$ visits an unseen place, the MaxAPs of $\cV_1$ and $\cV_2$ are small, we are uncertain about the matching result. Once $\cQ$ is finished, the new place discovered by $\cQ$ is added to the map to expand the coverage area. In addition, since $\c{Q}$ is matched against both $\cV_1$ and $\cV_2$, we can connect $\cV_1$ and $\cV_2$.}
	\label{fig:HMM_demo_and_new_connection}
	\vspace{-0.2cm}
\end{figure*}

\subsection{Related works}



PR has been addressed extensively in literature~\cite{lowry2016survey}. Traditionally, it has been posed as an image retrieval problem using local features aggregated via a BoW representation~\cite{FABMAP,BBOW,FABMAP2}. Feature-based methods fail to match correctly under appearance change. To address appearance change, SeqSLAM \cite{SeqSLAM} proposed to match statistics of the current image sequence to a sequence of images seen in the past, exploiting the temporal relationship. Recent methods have also looked at appearance transfer \cite{porav2018adversarial}\cite{latif2018addressing} to explicitly deal with appearance change.

The method closest in spirit to ours is~\cite{churchill2013experience}, who maintain multiple visual ``experiences'' of a particular location based on localization failures. In their work, successful localization leads to discarding data, and they depend extensively on visual odometry (VO), which can be a failure point. In contrast to~\cite{churchill2013experience}, our method does not rely on VO; only image sequences are required. Also, we update appearance in both successful and unsuccessful (new place) localization episodes, thus gaining robustness against appearance variations of the same place. Our method also has a novel mechanism for map compression leading to scalable inference.

A related problem is that of visual localization (VL): inferring the 6 DoF pose of the camera, given an image.
Given a model of the environment, PnP \cite {lepetit2009epnp} based solutions compute the pose using 2D-3D correspondences \cite{sattler2017efficient},
which becomes difficult both at large scale and under appearance change \cite{torii247Place}.
Some methods address the issue with creating a model locally using SfM against which query images are localized \cite{sattler2018benchmarking}. 
Given the ground truth poses and the corresponding images, VL can also be formulated as an image to pose regression problem, solving simultaneously the retrieval and pose estimation.
Recently, PoseNet \cite{kendall2015posenet} used a Convolution Neural Network (CNN) to learn this mapping,
with further improvements using LSTMs to address overfitting \cite{walch2017image}, uncertainty prediction \cite{kendall2016modelling} and inclusion of geometric constraints \cite{kendall2017geometric}.
MapNet \cite{brahmbhatt2018mapnet} showed that a representation of the map can be learned as a network and then used for VL. 
A downside of deep learning based methods is their high-computational cost to train/update.

Hidden Markov Models (HMMs) \cite{thrun2005probabilistic,russell2016artificial} have been used extensively for robot localization in indoor spaces~\cite{kosecka2004vision,aycard1997place,thrun1998probabilistic}. Hansen et al.~\cite{hansen2014visual} use HMM for outdoor scene, but they must maintain a similarity matrix between database and query sequences, which is unscalable when data is accumulated continuously. Therefore, we are one of the first to apply HMMs to large urban-scale PR, which requires significant innovation such as a novel efficient-to-evaluate observation model based on fast image retrieval (Sec.~\ref{sec:observation}). In addition, our method explicitly deals with temporal reasoning (Sec.~\ref{sec:transition}), which helps to combat the confusion from perceptual aliasing problem \cite{savinov2018semi}. Note also that our main contributions are in Sec.~\ref{sec:compress}, which tackles PR on a continuously growing dataset $\cD$.

\section{Map representation}\label{sec:map}


When navigating on a road network, the motion of the vehicle is restricted to the roads, and the heading of the vehicle is also constrained by the traffic direction. Hence, the variation in pose of the camera is relatively low \cite{sattler2018benchmarking,rubino2018practical}. 

The above motivates us to represent a road network as a graph $\mathcal{G} = (\mathcal{N},\cE)$,  which we also call the ``map". The set of nodes $\cN$ is simply the set of all images in $\cD$. To reduce clutter, we ``unroll" the image indices in $\cD$ by converting an $(i,j)$ index to a single number $k = N_1 + N_2 + \cdots + N_{i-1} + j$, hence the set of nodes are
\begin{align}\label{eq:initN}
\mathcal{N} = \{ 1, \dots, K \},
\end{align}
where $K = \sum_{i=1}^{M} N_i$ is the total number of images. We call an index $k \in \cN$ a ``place" on the map.

We also maintain a corpus $\cC$ that stores the images observed at each place. For now, the corpus simply contains
\begin{align}\label{eq:initC}
\cC(k) = \{ I_k \}, \;\;\;\; k = 1,\dots,K,
\end{align}
at each cell $\cC(k)$. Later in Sec.~\ref{sec:compress}, we will incrementally append images to $\cC$ as the video datatset $\cD$ grows.

In $\cG$, the set of edges $\cE$ connect images that overlap in their FOVs, i.e., $\langle k_1, k_2 \rangle$ is an edge in $\cE$ if
\begin{align}\label{eq:edges}
\exists I \in \cC(k_1)~\text{and}~\exists I^\prime \in \cC(k_2)~\text{such that}~I, I^\prime~~\text{overlap}.
\end{align}
Note that two images can overlap even if they derive from different videos and/or conditions. The edges are weighted by probabilities of transitioning between places, i.e.,
\begin{align}\label{eq:edges2}
w(\langle k_1, k_2 \rangle) =  P( k_2  \mid k_1) = P( k_1  \mid k_2),
\end{align}
for a vehicle that traverses the road network. Trivially,
\begin{align}
\langle k_1, k_2 \rangle \notin \cE~~\text{iff}~~P(k_2  \mid k_1) =  P( k_1  \mid k_2) = 0.
\end{align}
It is also clear from~\eqref{eq:edges2} that $\cG$ is undirected. Concrete definition of the transition probability will be given in Sec.~\ref{sec:compress}. First, Sec.~\ref{sec:hmm} discusses PR of $\cQ$ given a fixed $\cD$ and map.

\section{Place recognition using HMM}\label{sec:hmm}

To perform PR on $\cQ = \{Q_1,\dots,Q_T\}$ against a fixed map $\cG = (\cN,\cE)$ and corpus $\cC$, we model $\cQ$ using a HMM~\cite{russell2016artificial}. We regard each image $Q_t$ to be a noisy observation (image) of an latent \emph{place state} $s_t$, where $s_t \in \cN$. The main reason for using HMM for PR is to exploit the temporal order of the images in $\cQ$, and the high correlation between time and place due to the restricted motion (Sec.~\ref{sec:map}).


To assign a value to $s_t$, we estimate the belief
\begin{align}\label{eq:posterior}
P(s_t \mid Q_{1:t}), \;\;\;\; s_t \in \cN,
\end{align}
where $Q_{1:t}$ is a shorthand for $\{Q_1,\dots,Q_t \}$. Note that the belief is a probability mass function, hence
\begin{align}
\sum_{s_t \in \cN} P(s_t \mid Q_{1:t}) = 1.
\end{align}
Based on the structure of the HMM, the belief~\eqref{eq:posterior} can be recursively defined using Bayes' rule as
\begin{align}\label{eq:bayes}
\begin{aligned}
P( s_t | Q_{1:t}) = &\eta P(Q_t | s_t) \ast \\
&\sum_{s_{t-1} \in \cN} P(s_t | s_{t-1}) P(s_{t-1} | Q_{1:t-1}),
\end{aligned}
\end{align}
where $P(Q_t | s_t)$ is the observation model, $P(s_t | s_{t-1})$ is the state transition model, and $P(s_{t-1} | Q_{1:t-1})$ is the prior (the belief at the previous time step) \cite{russell2016artificial}. The scalar $\eta$ is a normalizing constant to ensure that the belief sums to $1$.

If we have the belief $P(s_t \mid Q_{1:t})$ at time step $t$, we can perform PR on $Q_t$ by assigning
\begin{align}\label{eq:map}
s_t^\ast = \argmax_{s_t \in \cN}~P(s_t \mid Q_{1:t})
\end{align}
as the place estimate of $Q_t$. Deciding the target state in this manner is called maximum \emph{a posteriori} (MaxAP) estimation. See Fig.~\ref{fig:HMM_demo_and_new_connection} for an illustration of PR using HMM.

\subsection{State transition model}\label{sec:transition}

The state transition model $P(s_t | s_{t-1})$ gives the probability of moving to place $s_{t}$, given that the vehicle was at place $s_{t-1}$ in the previous time step. The transition probability is simply given by the edge weights in $\cG$, i.e.,
\begin{align}
P(s_t = k_2 | s_{t-1} = k_1) = w(\langle k_1,k_2 \rangle).
\end{align}
Again, we defer the concrete definition of the transition probability to Sec.~\ref{sec:compress}. For now, the above is sufficient to continue our description of our HMM method.

\subsection{Observation model}\label{sec:observation}



Our observation model is based on image retrieval. Specifically, we use SIFT features~\cite{sift} and VLAD \cite{jegou2010aggregating} to represent every image. Priority search k-means tree \cite{flann} is used to index the database, but it is possible to use other indexing methods \cite{jegou2011product,douze2016polysemous,babenko2015tree}.

\vspace{-1em}
\paragraph{Image representation}

For every image $I_{k} \in \cC$, we seek a nonlinear function $\psi(I_k)$ that maps the image to a single high-dimensional vector. To do that, given a set of SIFT features densely extracted from image $I_k$: $\bX_{k} = \{x^h_k\} \in \mathbb{R}^{d \times H_k}$, where $H_k$ is the number of SIFT features of image $I_k$. K-means is used to build a codebook $B = \{b_m \in \mathbb{R}^d \, | \, m = 1, ...,\mathtt{M} \}$, where $\mathtt{M}$ is the size of codebook. The VLAD embedding function is defined as: 
\begin{align}
\phi(x_k) = [..., 0, x^h_k - b_m, 0, ...] \in \mathbb{R}^D
\end{align}
where, $b_m$ is the nearest visual word of feature vector $x^h_k$. To obtain a single vector, we employ sum aggregation:
\begin{align}
\psi(I_k) = \displaystyle\sum_{i=1}^{H_k} \phi(x_k)
\end{align}
To reduce the impact of background features (e.g., trees, roads, sky) within the vector $\psi(I_k)$, we adopt rotation and normalization (RN) \cite{jegou2014triangulation}, followed by $L$-2 normalization. In particular, we use PCA to project $\psi(I_k)$ from $D$ to $D'$, where $D' < D$. In our experiment, we set $D'= 4,096$. Power-law normalization is then applied on rotated data: 
\begin{align}
\psi(I_k) := |\psi(I_k)|^\alpha sign(\psi(I_k))
\end{align}
where, we set $\alpha = 0.5$. 

Note that different from DenseVLAD \cite{torii2015densevlad} which uses whitening for post-processing, performing power-law normalization on rotated data is more stable.

\vspace{-1.25em}
\paragraph{Computing likelihood}

We adopt priority search k-means tree \cite{flann} to index every image $I_k \in \cC$. The idea is to partition all data points $\psi(I_k)$ into $\bK$ clusters by using K-means, then recursively partitioning the points in each cluster. For each query $Q_t$, we find a set of $L$-nearest neighbor $\bL(Q_t)$. Specifically, $Q_t$ is mapped to vector $\psi(Q_t)$. To search, we propagate down the tree at each cluster by comparing $\psi(Q_t)$ to $\bK$ cluster centers and selecting the nearest one. 

The likelihood $P(Q_t|s_t)$ is calculated as follows:
\vspace{-\topsep}
\begin{itemize}
	\setlength{\parskip}{0pt}
	\setlength{\itemsep}{0pt plus 1pt}
	\item Initialize $P(Q_t|s_t=k) = e^{\frac{-\beta}{\sigma}}$, $\forall k \in \cN$, where, we set $\beta = 2.5$ and $\sigma = 0.3$ in our experiment.
	
	\item For each $I_k \in \bL(Q_t)$
	
	\vspace{-\topsep}
	\begin{itemize}
		\setlength{\parskip}{0pt}
		\setlength{\itemsep}{0pt plus 1pt}
		\item Find node $\hat{k} = \cC^{-1}(I_k)$, where $\cC^{-1}$ is the inverse of corpus $\cC$, which finds node $\hat{k}$ storing $I_k$.
		
		\item Calculate the probability: $\hat{p} = e^{\frac{-dist(Q_t, I_k)}{\sigma} }$, where $dist$ is the distance between $Q_t$ and $ I_k$.
		
		\item If $\hat{p} > P(Q_t|s_t=\hat{k})$, then $P(Q_t|s_t=\hat{k}) = \hat{p}$.
	\end{itemize}	
\end{itemize}

\subsection{Inference using matrix computations}

%

The state transition model can be stored in a $K \times K$ matrix $\bE$ called the \emph{transition matrix}, where the element at the $k_1$-th row and $k_2$-th column of $\bE$ is
\begin{align}
\bE(k_1,k_2) = P(s_t = k_2 \mid s_{t-1} = k_1).
\end{align}
Hence, $\bE$ is also the weighted adjacency matrix of graph $\cG$.  Also, each row of $\bE$ sums to one. The observation model can be encoded in a $K \times K$ diagonal matrix $\bO_t$, where
\begin{align}
\mathbf{O}_{t}(k,k) = P( Q_t \mid s_t = k ).
\end{align}
If the belief and prior are represented as vectors $\bp_{t}, \bp_{t-1} \in \mathbb{R}^K$ respectively, operation~\eqref{eq:bayes} can be summarized as
\begin{align}\label{eq:inference}
\bp_{t} = \eta \bO_{t} \bE^T \bp_{t-1},
\end{align}
where $\bp_0$ corresponds to uniform distribution. From this, it can be seen that the cost of PR is $\mathcal{O}(K^2)$.

\vspace{-1em}
\paragraph{Computational cost}

Note that $\bE$ is a very sparse matrix, due to the topology of the graph $\cG$ which mirrors the road network; see Fig.~\ref{fig:mapillary_update_transition_matrix_example} for an example $\bE$. Thus, if we assume that the max number of non-zero values per row in $\bE$ is $r$, the complexity for computing $\bp_t$ is O$(rK)$.

Nonetheless, in the targeted scenario (Sec.~\ref{sec:setting}), $\cD$ can grow unboundedly. Thus it is vital to avoid a proportional increase in $\bE$ so that the cost of PR can be maintained.

\section{Scalable place recognition based on HMM}\label{sec:compress}

\begin{figure*}[h]
	\centering
	\mbox
	{		
		\subfloat[][Matching]
		{
			\includegraphics[width=0.47\textwidth]{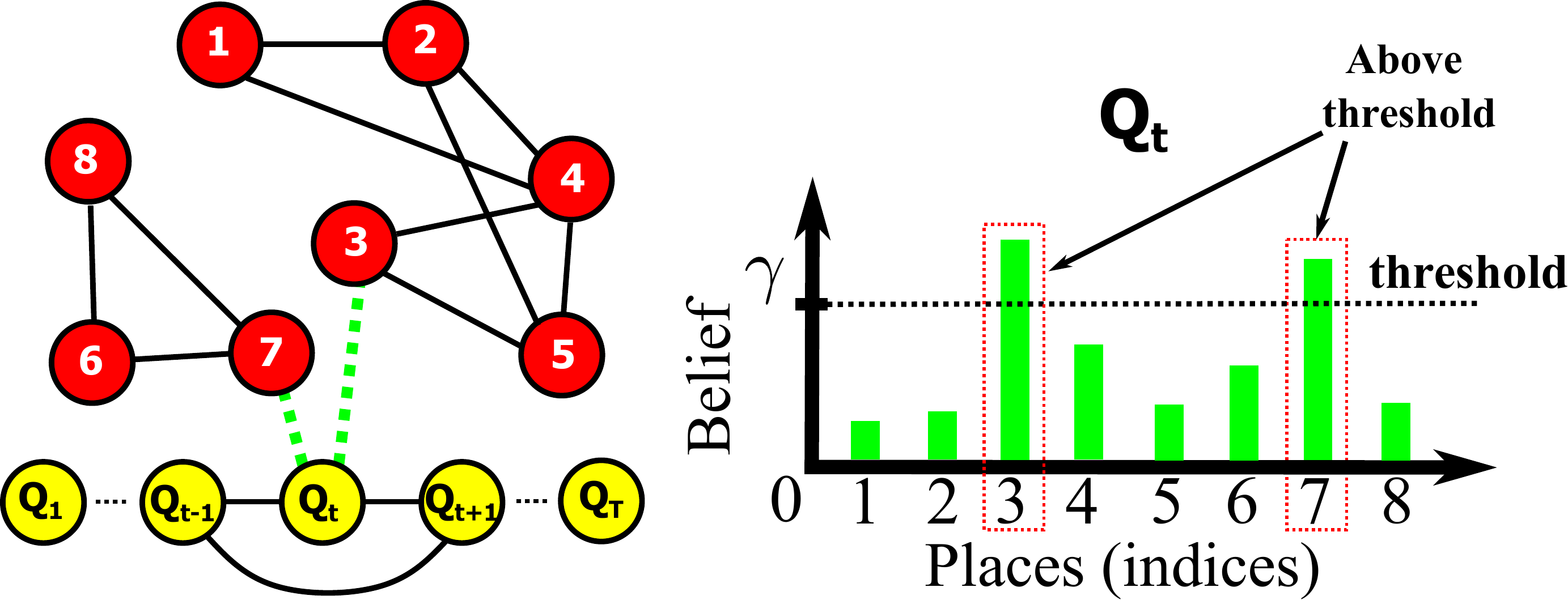} 
			\label{fig:scalable_pc_matching}
		}

		\hspace{0.3cm}
		\subfloat[][Culling]
		{
			\includegraphics[width=0.2\textwidth]{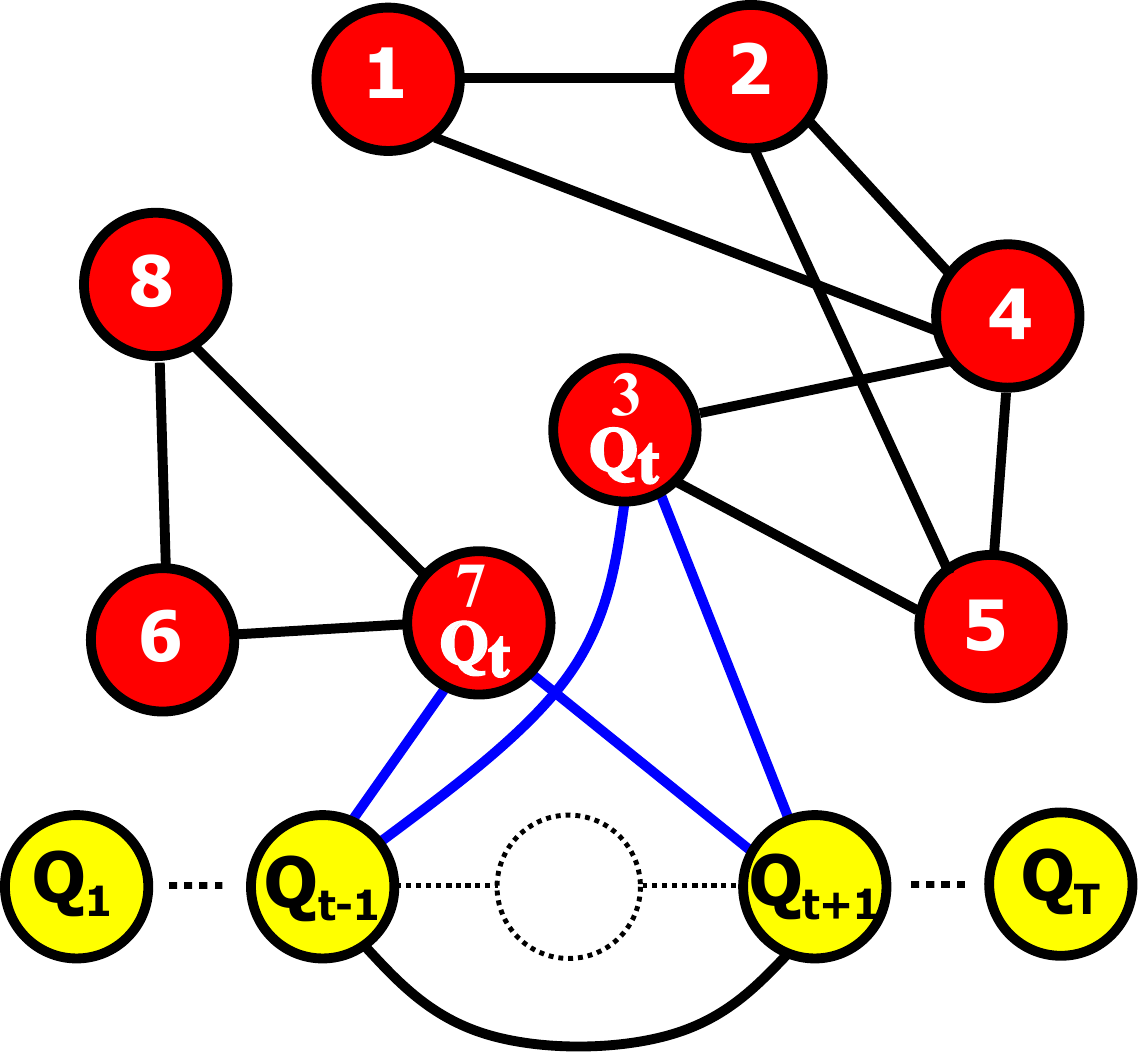} 
			\label{fig:scalable_pc_update}
		}
		
		\hspace{0.3cm}
		
		\subfloat[][Combining]
		{
			\includegraphics[width=0.2\textwidth]{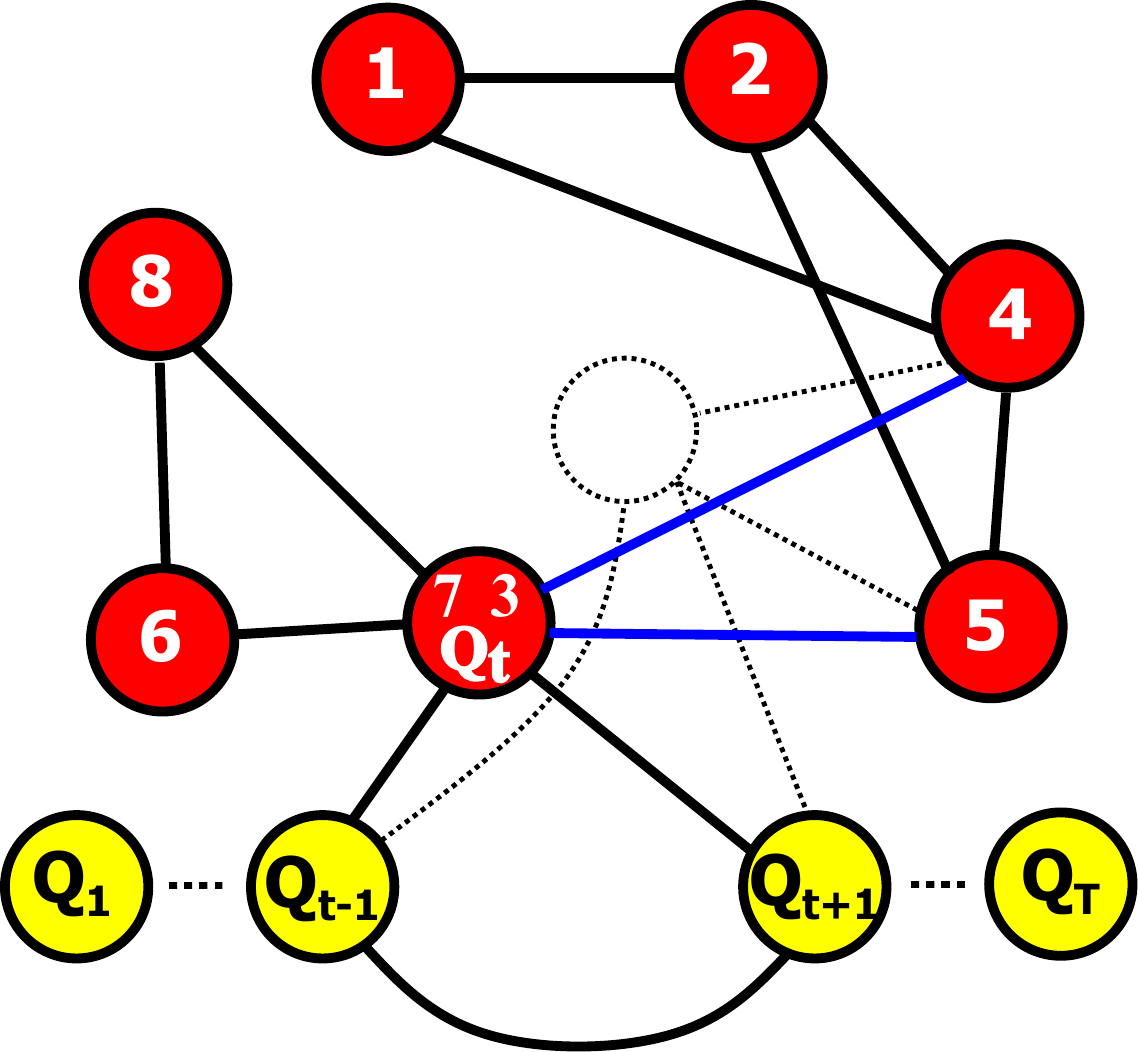} 
			\label{fig:scalable_pc_compress}
			}
	}

	\caption{An overview of our idea for scalable place recognition. Graph $\cG = \cG_1 \cup \cG_2$, where $\cG_1 = \{1,2,3,4,5\}$ and $\cG_2 = \{6,7,8\}$ are disjoint sub-graphs. Query video $\cQ= \{Q_1, ..., Q_T \}$ is matched against $\cG$. Figure \ref{fig:scalable_pc_matching}: $Q_t$ is matched with node $k=3$ and $7$ (dashed green lines), due to $\bp_t(3)$, $\bp_t(7) > \gamma$. Figure \ref{fig:scalable_pc_update}: $Q_t$ is added to node $3$ and $7$, new edges are created (blue lines) to maintain the connections between $Q_{t-1}$, $Q_{t+1}$ and $Q_t$. Figure \ref{fig:scalable_pc_compress}: Node $3$ and $7$ are combined. New edges are generated (blue lines) to maintain the connections within the graph. Note that after matching query $\cQ$ against $\cG$, our proposed culling and combining methods connect two disjoint sub-graphs $\cG_1$ and $\cG_2$ together.}
	\label{fig:scalable_pc}
\end{figure*}

In this section, we describe a novel method that incrementally builds and compresses $\cG$ for a video dataset $\cD$ that grows continuously due to the addition of new query videos.

We emphasize again that the proposed technique functions without using GNSS positioning or visual odometry.

\subsection{Map intialization}\label{sec:mapinit}

Given a dataset $\cD$ with one video $\cV_1 = \{ I_{1,j} \}^{N_1}_{j=1} \equiv \{ I_{k} \}^{K}_{k=1}$, we initialize $\cN$ and $\cC$ as per~\eqref{eq:initN} and~\eqref{eq:initC}. The edges $\cE$ (specifically, the edge weights) are initialized as
\begin{align*}\label{eq:initweights}
w(\langle k_1,k_2 \rangle) = \begin{cases} 0 & \text{if}~|k_1 - k_2| > W, \\ \alpha\exp\left( - \frac{|k_1 - k_2|^2}{\delta^2} \right) & \text{otherwise}, \end{cases}
\end{align*}
where $\alpha$ is a normalization constant. The edges connect frames that are $\le W$ time steps apart with weights based on a Gaussian on the step distances. The choice of $W$ can be based on the maximum velocity of a vehicle.

Note that this simple way of creating edges will ignore complex trajectories (e.g., loops). However, the subsequent steps will rectify this issue by connecting similar places.

\subsection{Map update and compression}\label{sec:mapupdate}

Let $\cD = \{ \cV_i \}^{M}_{i=1}$ be the current dataset with map $\cG = (\cN,\cE)$ and corpus $\cC$. Given a query video $\cQ = \{ Q_t \}^{T}_{t=1}$, using our method in Sec.~\ref{sec:hmm} we perform PR on $\cQ$ based on $\cG$. This produces a belief vector $\bp_t$~\eqref{eq:inference} for all $t$.

We now wish to append $\cQ$ to $\cD$, and update $\cG$ to maintain computational scalability of future PR queries. First, create a subgraph $\cG^\prime = (\cN^\prime, \cE^\prime)$ for $\cQ$, where
\begin{align}
\cN^\prime = \{ K+1,K+2, \dots, K+T \},
\end{align}
(recall that there are a total of $K$ places in $\cG$), and $\cE^\prime$ simply follows Sec.~\ref{sec:mapinit} for $\cQ$. 

In preparation for map compression, we first concatenate the graphs and extend the corpus
\begin{align}
\cN = \cN \cup \cN^\prime,~\cE = \cE \cup \cE^\prime,~\text{and}~\cC(K+t) = \{ Q_t \}
\end{align}
for $t = 1,\dots,T$. There are two main subsequent steps: culling new places, and combining old places. 

\vspace{-1em}
\paragraph{Culling new places}

For each $t$, construct
\begin{align}
\cM(t) = \{ k \in \{1,\dots,K\} \mid \bp_t(k) \ge \gamma \},
\end{align}
where $\gamma$ with $0 \le \gamma \le 1$ is a threshold on the belief. There are two possibilities:

\vspace{-\topsep}
\begin{itemize}
	\setlength{\parskip}{0pt}
	\setlength{\itemsep}{0pt plus 1pt}
	\item If $\cM(t) = \emptyset$, then $Q_t$ is the image of a new (unseen before) place since the PR did not match a dataset image to $Q_t$ with sufficient confidence. No culling is done.
	\item If $\cM(t) \ne \emptyset$, then for each $k_1 \in \cM(t)$,

	\vspace{-\topsep}
	\begin{itemize}
		\setlength{\parskip}{0pt}
		\setlength{\itemsep}{0pt plus 1pt}
		\item For each $k_2$ such that $\langle K+t ,k_2 \rangle \in \cE$:
		
		\begin{itemize}

			\item Create new edge $\langle k_1, k_2 \rangle$ with weight $w(\langle k_1, k_2 \rangle) = w(\langle K+t, k_2 \rangle)$.
			
			\item Delete edge $\langle K+t ,k_2 \rangle$ from $\cE$.
		\end{itemize}
		
		\item $\cC(k_1) = \cC(k_1) \cup \cC(K+t)$.
	\end{itemize}
\end{itemize}

\vspace{-0.3cm}
Once the above is done for all $t$, for those $t$ where $\cM(t) \ne \emptyset$, we delete the node $K+t$ in $\cN$ and cell $\cC(K+t)$ in $\cC$, both with
the requisite adjustment in the remaining indices. See Figs.~\ref{fig:scalable_pc_matching} and \ref{fig:scalable_pc_update} for an illustration of culling.

\vspace{-1em}
\paragraph{Combining old places}

Performing PR on $\cQ$ also provides a chance to connect places in $\cG$ that were not previously connected. For example, two dataset videos $\cV_1$ and $\cV_2$ could have traversed a common subpath under very different conditions. If $\cQ$ travels through the subpath under a condition that is simultaneously close to the conditions of $\cV_1$ and $\cV_2$, this can be exploited for compression.

To this end, for each $t$ where $\cM(t)$ is non-empty,
\vspace{-\topsep}
\begin{itemize}
	\setlength{\parskip}{0pt}
	\setlength{\itemsep}{0pt plus 1pt}
	\item $k_1 = \min \cM(t)$.
	\item For each $k_2 \in \cM(t)$ where $k_2 \ne k_1$ and $\langle k_1, k_2 \rangle \notin \cE$:
	
	\vspace{-\topsep}
	\begin{itemize}
		\item For each $k_3$ such that $\langle k_2, k_3 \rangle \in \cE$, $\langle k_1, k_3 \rangle \notin \cE$:
		
		\vspace{-\topsep}
		\begin{itemize}
			\item Create  edge $\langle k_1, k_3 \rangle$ with weight 
			$w(\langle k_1, k_3 \rangle) = w(\langle k_2, k_3 \rangle)$.
			\item Delete edge $\langle k_2 ,k_3 \rangle$ from $\cE$.
		\end{itemize}
		\item $\cC(k_1) = \cC(k_1) \cup \cC(k_2)$.
	\end{itemize}
\end{itemize}
\vspace{-0.2cm}
\noindent Again, once the above is done for all $t$ for which $\cM(t) \ne \emptyset$, we remove all unconnected nodes from $\cG$ and delete the relevant cells in $\cC$, with the corresponding index adjustments. Figs.~\ref{fig:scalable_pc_compress}, \ref{fig:HMM_new_connection_1} and \ref{fig:HMM_new_connection_2} illustrate this combination step.

\subsection{Updating the observation model}\label{sec:flannupdate}


When $\cQ$ is appended to the dataset, i.e., $\cD = \cD \cup \cQ$, all vector $\psi(Q_t)$ need to be indexed to the k-means tree. In particular, we find the nearest leaf node that $\psi(Q_t)$ belongs to. Assume the tree is balanced, the height of tree is $(log \, N / log \, \bK)$, where $N = \sum N_i$, thus each $\psi(Q_t)$ needs to check $(log \, N / log \, \bK)$ internal nodes and one leaf node. In each node, it needs to find the closest cluster center by computing distances to all centers, the complexity of which is $O(\bK D')$. Therefore, the cost for adding the query video $\cQ$ is $O\begin{pmatrix}T\bK D' (log \, N / log \, \bK)\end{pmatrix}$, where $T = |\cQ|$. Assume it is a complete tree, every leaf node contains $\bK$ points, thus it has $N/\bK$ leaf nodes. For each point $\psi(Q_t)$, instead of exhaustedly scanning $N/\bK$ leaf nodes, it only needs to check $log \, N / log \, \bK$ nodes. Hence, it is a scalable operation.

\subsection{Overall algorithm}

\begin{figure*}[h]
	
	\centering
	\includegraphics[width=0.9\textwidth]{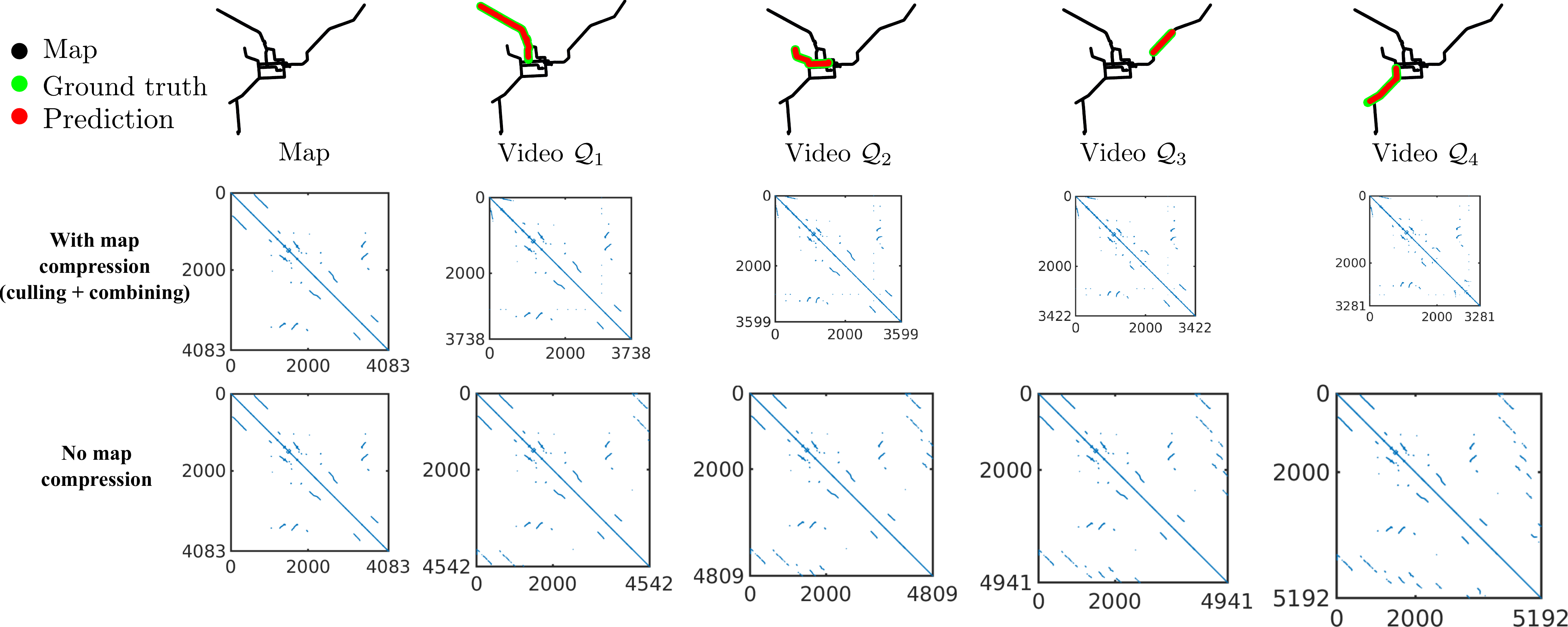}
	
	\vspace{-0.75em}
	\caption{Illustrating map maintenance w and w/o compression. After each query video $\cQ$ finishes, we compress the map by culling known places in $\cQ$ and combining old places on the map which represent the same place. Thus, the size of transition matrix is shrunk gradually. In contrast, if compression is not conducted, the size of transition matrix will continue increasing.}
	\label{fig:mapillary_update_transition_matrix_example}
	\vspace{-0.2cm}
\end{figure*}

Algorithm~\ref{alg:overall} summarizes the proposed scalable method for PR. A crucial benefit of performing PR with our method is that map $\cG$ does not grow unboundedly with the inclusion of new videos. Moreover, the map update technique is simple and efficient, which permits it to be conducted for every new video addition. This enables scalable PR on an ever growing video dataset. In Sec.~\ref{sec:comparison_sota}, we will compare our technique with state-of-the-art PR methods.




\section{Experiments}
\
We use a dataset sourced from Mapillary \cite{neuhold2017mapillary} which consists of street-level geo-tagged imagery; see supplementary material for examples.
Benchmarking was carried out on the Oxford RobotCar \cite{maddern20171},
from which we use 8 different sequences along the same route; details are provided in supplementary material, and the sequences are abbreviated as Seq-1 to Seq-8. The initial database $\cD$ is populated with
Seq-1 and Seq-2 from the Oxford RobotCar dataset. Seq-3 to Seq-8 are then sequentially used as the query videos. 
To report the 6-DoF pose for a query image, we inherit the pose of the image matched using the MaxAP estimation.
Following \cite{sattler2018benchmarking}, the translation error is computed as the Euclidean distance $||c_{est} - c_{gt}||^2$. Orientation errors $|\theta|$, measured in degree, is the angular difference $2 \, cos(|\theta|) = trace(R_{gt}^{-1} \, R_{est}) -1$ between estimated and ground truth camera rotation matrices $R_{est}$ and $R_{gt}$. Following \cite{kendall2015posenet,kendall2017geometric,brahmbhatt2018mapnet,wang2018dels}, we compare mean and median errors.

\vspace{-1.1em}
\paragraph{Performance with and without updating the database}

\begin{algorithm}[ht]\centering
	\begin{algorithmic}[1]
		\REQUIRE Threshold $W$ for transition probability, threshold $\gamma$ for PR, initial dataset $\cD = \{ \cV_1 \}$ with one video.
		\STATE Initialize map $\cG = (\cN,\cE)$ and corpus $\cC$ (Sec.~\ref{sec:mapinit}).
		\STATE Create observation model (Sec.~\ref{sec:observation})
		\WHILE{there is a new query video $\cQ$}
		\STATE Perform PR on $\cQ$ using map $\cG$, then append $\cQ$ to $\cD$.\vspace{-1em}
		\STATE Create subgraph $\cG^\prime$ for $\cQ$ (Sec.~\ref{sec:mapupdate}).
		\STATE Concatenate $\cG^\prime$ to $\cG$, extend $\cC$ with $\cQ$ (Sec.~\ref{sec:mapupdate}).
		\STATE Reduce $\cG$ by culling new places (Sec.~\ref{sec:mapupdate}).
		\STATE Reduce $\cG$ by combining old places (Sec.~\ref{sec:mapupdate}).
		\STATE Update observation model (Sec.~\ref{sec:flannupdate}).
		\ENDWHILE
		\RETURN Dataset $\cD$ with map $\cG$ and corpus $\cC$.
	\end{algorithmic}
	\caption{Scalable algorithm for large-scale PR.}
	\label{alg:overall}
\end{algorithm}

\begin{table}
	
	\small
	\begin{tabular}{c}
		\begin{tabular}{l |c | c | c }
			 & No update & Cull & Cull+combine \\
			\hline
			\hline
			Seq-3 & \multicolumn{3}{c}{6.59m, 3.28$^\circ$}  \\
			\hline
			Seq-4 & 7.42m, 4.64$^\circ$ & \textbf{5.80m}, 3.24$^\circ$ & 6.01m, \textbf{3.11$^\circ$} \\
			\hline
			Seq-5 & 16.21m, 5.97$^\circ$ & \textbf{15.07m}, \textbf{5.89$^\circ$} & 15.88m, 5.91$^\circ$\\
			\hline
			Seq-6 & 26.02m, 9.02$^\circ$ & \textbf{18.88m}, \textbf{6.24$^\circ$} & 19.28m, 6.28$^\circ$ \\
			\hline
			Seq-7 & 31.83m, 17.99$^\circ$ & 30.06m, 17.12$^\circ$ & \textbf{30.03m}, \textbf{17.05$^\circ$} \\
			\hline
			Seq-8 & 25.62m, 22.38$^\circ$ & 24.28m, 21.99$^\circ$ & \textbf{24.26m}, \textbf{21.54$^\circ$}\\
			\hline

		\end{tabular}
		\\
\\

			\begin{tabular}{l |c | c | c }
				& No update & Cull & Cull+combine \\
				\hline
				\hline
				Seq-3 & \multicolumn{3}{c}{6.06m, 1.65$^\circ$}  \\
				\hline
				Seq-4 & 5.80m, 1.40$^\circ$ & \textbf{5.54m}, 1.39$^\circ$ & 5.65m, \textbf{1.33$^\circ$} \\
				\hline
				Seq-5 & 13.70m, 1.56$^\circ$ & 13.12m, \textbf{1.52$^\circ$} & \textbf{13.05m}, 1.55$^\circ$\\
				\hline
				Seq-6 & 6.65m, 1.87$^\circ$ & \textbf{5.76m}, \textbf{1.75$^\circ$} &  6.60m, 1.85$^\circ$\\
				\hline
				Seq-7 & 13.58m, 3.52$^\circ$ & 11.80m, 2.81$^\circ$ & \textbf{10.87m}, \textbf{2.60$^\circ$} \\
				\hline
				Seq-8 & 13.28m, 4.93$^\circ$ & \textbf{7.13m}, \textbf{2.31$^\circ$} & 7.15m, 2.47$^\circ$\\
				\hline

			\end{tabular}
		
	\end{tabular}
	
	\caption{Comparison between 3 different settings of our technique. Mean (top) and median (bottom) errors of 6-DoF pose on Oxford RobotCar are reported.}
	\label{tab:benefits_update}
	\vspace{-2em}
\end{table}

We investigate the effects of updating database on localization accuracy and inference time. 
After each query sequence finishes, we consider three strategies: 
\begin{inparaenum}[i)] 
	\item \texttt{No update}: $\cD$ always contains just the initial 2 sequences,
	\item \texttt{Cull}: Update $\cD$ with the query and perform culling, and 
	\item \texttt{Cull+Combine}: Full update with both culling and combining nodes.
\end{inparaenum}
Mean and median 6-DoF pose errors are reported in Table \ref{tab:benefits_update}. 
In general, \texttt{Cull} improves the localization accuracy over \texttt{No update},
since culling adds appearance variation to the map. 
In fact, there are several cases, in which \texttt{Cull+Combine} produces better results over \texttt{Cull}. 
This is because we consolidate useful information in the map (combining nodes which represent the same place), 
and also enrich the map topology (connecting nodes close to each other through culling).
Inference times per query with different update strategies are given in Table \ref{tab:inference_time}. 
Without updating, the inference time is stable at ($\sim 4$ms/query) between sequences, since the size of graph and the database do not change. 
In contrast, culling operation increases the inference time by about $1$ms/query, 
and \texttt{Cull+Combine} makes it comparable to the \texttt{No update} case.
This shows that the proposed method is able to compress the database to an extent that the query time after assimilation of new information remains comparable to the case of not updating the database at all.

\begin{table}
	\centering
	\small
	\begin{tabular}{ l | c | c | c }
		\hline
		Sequences & No update & Cull & Cull+Combine  \\
		\hline
		\hline
		
		Seq-3 & \multicolumn{3}{c}{4.03} \\
		\hline
		Seq-4 & \textbf{4.56} & 5.05  &  4.82\\
		\hline
		Seq-5 & \textbf{4.24} & 5.06 & 4.87 \\
		\hline
		Seq-6 & 3.81 & 4.03 &  \textbf{3.72} \\
		\hline
		Seq-7 & 3.82 & 4.18 & \textbf{3.78} \\
		\hline
		Seq-8 & 3.77 & 3.91 & \textbf{3.68}\\
		\hline
	\end{tabular}
	\caption{Inference time (ms) on Oxford RobotCar. 
		\texttt{Cull+Combine} has comparable inference time while giving better accuracy (see Table \ref{tab:benefits_update}) over \texttt{No update}.}
	\label{tab:inference_time}
	\vspace{-1em}
\end{table}

\begin{table}
	\centering
	\small
	\begin{tabular}{ m{1.5cm} | c | c | m{1cm} }
		\hline
		Training sequences & VidLoc & MapNet & Our method \\
		\hline
		\hline
		Seq-1,2 & 14.1h & 11.6h & \textbf{98.9s} \\
		\hline
		Seq-3 & -  & 6.2h & \textbf{256.3s} \\
		\hline
		Seq-4 & - & 6.3h & \textbf{232.3s} \\
		\hline
		Seq-5 & - & 6.8h  & \textbf{155.1s }\\
		\hline
		Seq-6 & - & 5.7h & \textbf{176.5s} \\
		\hline
		Seq-7 & - & 6.0h  & \textbf{195.4s} \\
		\hline
	\end{tabular}
	\caption{Training/updating time on the Oxford RobotCar.}
	\label{tab:compare_sota_train_time}
\end{table}

\vspace{-1.25em}
\paragraph{Map maintenance and visiting unknown regions} \label{sec:experiment_map_maintain_visit_new_place}

\begin{figure*}
	\captionsetup[subfigure]{labelformat=empty}
	\centering
	\includegraphics[width=1\textwidth]{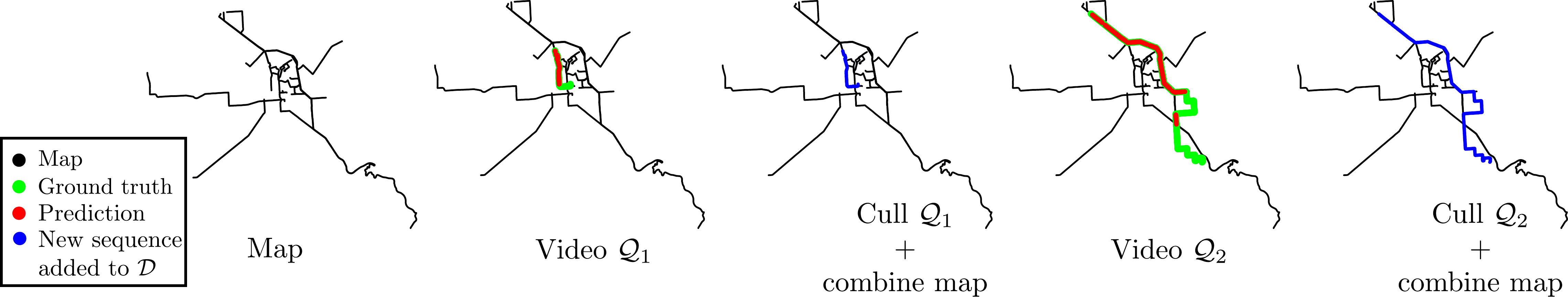}

	\caption{Expanding coverage by updating the map. Locations are plotted using ground-truth GPS for visualization only.}
	\label{fig:mapillary_visit_new_place}
\end{figure*}

Figure \ref{fig:mapillary_update_transition_matrix_example} shows the results on map maintenance with and without compression. Without compression, size of map $\cG$ (specifically, adjacency matrix $\bE$) grows continuously when appending a new query video $\cQ$. In contrast, using our compression scheme, known places in $\cQ$ are culled, and redundant nodes in $\cG$ (i.e., nodes representing a same place) are combined. As a result, the graph is compressed.

\begin{table*}
	
	\centering
	\small
	\begin{tabular}{c}
		\begin{tabular}{m{2.3cm} |c | c | c | c | c | c }
			
			Methods & Seq-3 & Seq-4 & Seq-5 & Seq-6 & Seq-7 & Seq-8 \\
			\hline
			\hline
			VidLoc & 38.86m, 9.34$^\circ$ & 38.29m, 8.47$^\circ$ & 36.05m, 6.81$^\circ$ & 51.09m, 10.75$^\circ$ & 54.70m, 18.74$^\circ$ & 47.64m, 23.21$^\circ$ \\
			\hline
			MapNet & \multirow{3}{*}{9.31m, 4.37$^\circ$} & 8.92m, 4.09$^\circ$ & 17.19m, \textbf{5.72$^\circ$}  & 26.31m, 9.78$^\circ$ & 33.68m, 18.04$^\circ$ & 26.55m, 21.97$^\circ$ \\
			
			\cline{1-1} \cline{3-7}
			
			MapNet (update+ retrain)  & & 8.71m, 3.31$^\circ$ & 18.44m, 6.94$^\circ$ & 28.69m, 10.02$^\circ$ & 36.68m, 19.34$^\circ$ & 29.64m, 22.86$^\circ$\\
			
			\hline
			
			Our method & \textbf{6.59m}, \textbf{3.28$^\circ$} & \textbf{6.01m}, \textbf{3.11$^\circ$} & \textbf{15.88m}, 5.91$^\circ$ & \textbf{19.28m}, \textbf{6.28$^\circ$} & \textbf{30.03m}, \textbf{17.05$^\circ$} & \textbf{24.26m}, \textbf{21.54$^\circ$} \\
			
			\hline
		\end{tabular}
		\\
		\\
		
		\begin{tabular}{m{2.3cm} |c | c | c | c | c | c }
			Methods & Seq-3 & Seq-4 & Seq-5 & Seq-6 & Seq-7 & Seq-8 \\
			\hline
			\hline
			VidLoc & 29.63m, \textbf{1.59$^\circ$} & 29.86m, 1.57$^\circ$ & 31.33m, 1.39$^\circ$ & 47.75m, \textbf{1.70$^\circ$} & 48.53m, 2.40$^\circ$ & 42.26m, \textbf{1.94$^\circ$} \\
			\hline
			MapNet & \multirow{3}{*}{\textbf{4.69m}, 1.67$^\circ$} & \textbf{4.53m}, 1.54$^\circ$ & 13.89m, \textbf{1.17$^\circ$}  & 8.69m, 2.42$^\circ$ & 12.49m, \textbf{1.71$^\circ$ }& 8.08m, 2.02$^\circ$ \\
			
			\cline{1-1} \cline{3-7}
			
			MapNet (update+ retrain)  & & 5.15m, 1.44$^\circ$ & 17.39m, 1.87$^\circ$ & 11.45m, 3.42$^\circ$ & 20.88m, 4.02$^\circ$ & 11.01m, 5.21$^\circ$ \\
			
			\hline
			
			Our method & 6.06m, 1.65$^\circ$ & 5.65m, \textbf{1.33$^\circ$} & \textbf{13.05m}, 1.55$^\circ$ & \textbf{6.60m}, 1.85$^\circ$ & \textbf{10.87m}, 2.60$^\circ$ & \textbf{7.15m}, 2.47$^\circ$ \\
			
			\hline
		\end{tabular}
	\end{tabular}
	
	\caption{Comparison between our method, MapNet and VidLoc. Mean (top) and median (bottom) 6-DoF pose errors on the Oxford RobotCar dataset are reported.}
	\label{tab:compare_sota_accuracy}
	
\end{table*}

Visiting unexplored area allows us to expand the coverage of our map, as we demonstrate using Mapillary data.
We set $\gamma = 0.3$, i.e., we only accept the query frame which has the MaxAP belief $\ge 0.3$. 
When the vehicle explores unknown roads, the probability of MaxAP is small and no localization results are accepted. 
Once the query sequence ends, the map coverage is also extended; see Fig.~\ref{fig:mapillary_visit_new_place}.

\vspace{-1.25em}
\paragraph{Comparison against state of the art} \label{sec:comparison_sota}

\begin{figure}
	
	\centering
	\includegraphics[width=0.48\textwidth]{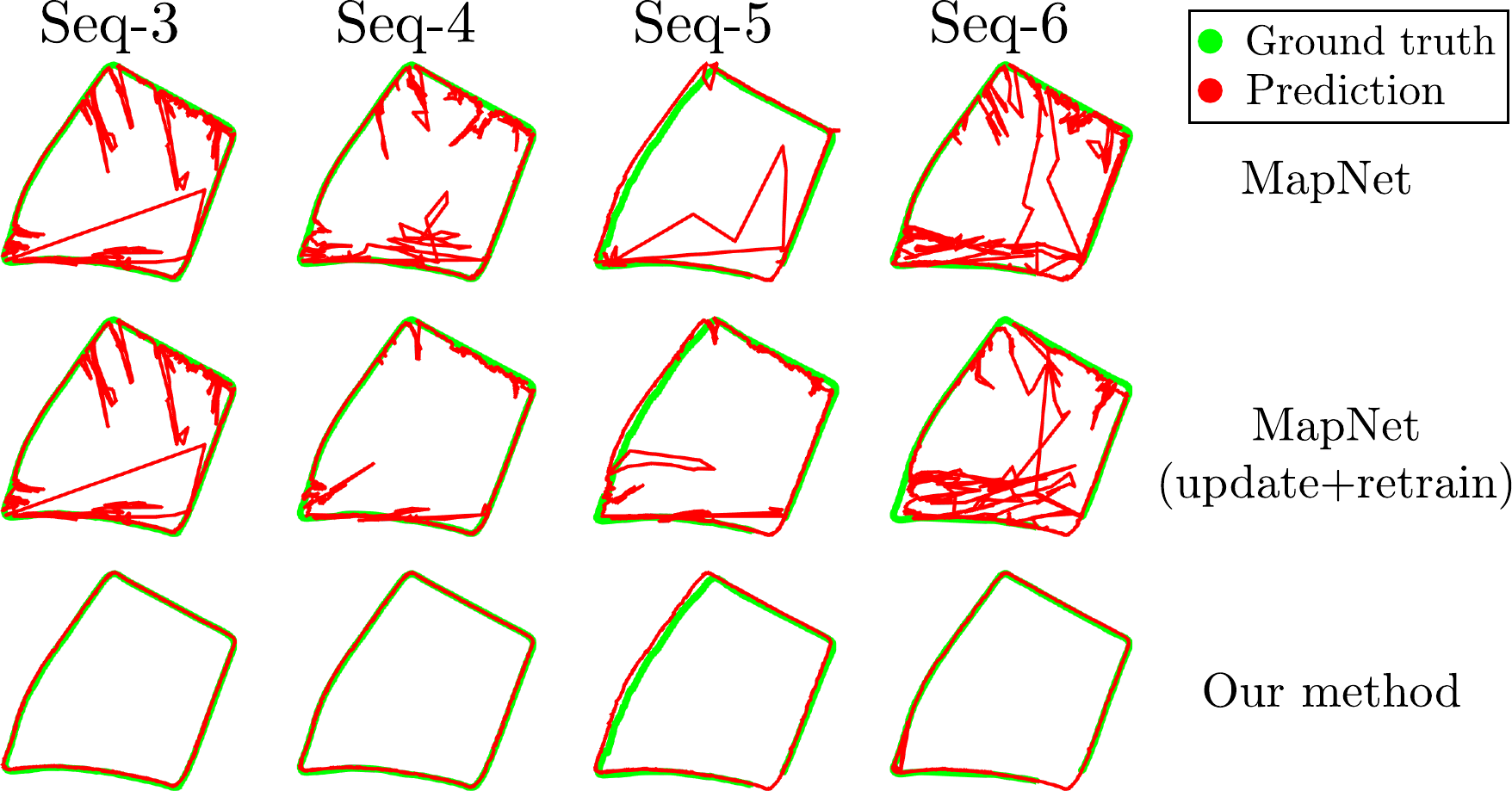}

	\caption{Qualitative results on the RobotCar dataset.}
	\label{fig:oxfordcar_quantitative}
	\vspace{-2em}
\end{figure}

Our method is compared against state-of-the-art localization methods: MapNet \cite{brahmbhatt2018mapnet} and VidLoc \cite{clark2017vidloc}.
We use the original authors' implementation of MapNet. VidLoc implementation from MapNet is used by the recommendation of VidLoc authors.
All parameters are set according to suggestion of authors.\footnote{Comparisons against \cite{churchill2013experience} are not presented due to the lack of publicly available  implementation.} 

For map updating in our method, \texttt{Cull+Combine} steps are used.
MapNet is retrained on the new query video with the ground truth from previous predictions. 
Since VidLoc does not produce sufficiently accurate predictions, we do not retrain the network for subsequent query videos.

Our method outperforms MapNet and VidLoc in terms of the mean errors (see Table \ref{tab:compare_sota_accuracy}), and also has a smoother predicted trajectory than MapNet (see Fig.~\ref{fig:oxfordcar_quantitative}). In addition, while our method improves localization accuracy after updating the database (See Table \ref{tab:benefits_update}), MapNet's results is worse after retraining (See Table \ref{tab:compare_sota_accuracy}). This is because MapNet is retrained on a noisy ground truth. However, though our method is qualitatively better than MapNet, differences in median error is not obvious: this shows that median error is not a good criterion for VL, since gross errors are ignored.

Note that our method mainly performs PR; here, comparisons to VL methods are to show that a correct PR paired with simple pose inheritance can outperform VL methods in presence of appearance change. The localization error of our method can likely be improved by performing SfM on a set of images corresponding to the highest belief.


Table \ref{tab:compare_sota_train_time} reports training/updating time for our method and MapNet and VidLoc. 
Particularly, for Seq-1 and Seq-2, our method needs around 1.65 minute to construct the k-means tree and build the graph,
while MapNet and VidLoc respectively require 11.6 and 14.1 hours for training. 
For updating a new query sequence, MapNet needs about 6 hours of retraining the network, whilst our method culls the database and combine graph nodes in less than 5 minutes. 
This makes our method more practical in a realistic scenario, in which the training data is acquired continuously.

\section{Conclusion}

This paper proposes a novel method for scalable place recognition, which is lightweight in both training and testing when the data is continuously accumulated to maintain all of the appearance variation for long-term place recognition. From the results, our algorithm shows significant potential towards achieving long-term autonomy.

{\small
\bibliographystyle{ieee}
\bibliography{egbib}
}

\section{Supplementary Material}

\begin{table*}
	\centering
	\small
	\begin{tabular}{|l|l|c|c|}
		\hline
		Abbreviation & Recorded & Condition & Sequence length \\
		\hline
		%
		Seq-1 & 26/06/2014, 09:24:58 & overcast & 3164 \\
		Seq-2 & 26/06/2014, 08:53:56 & overcast & 3040 \\
		Seq-3 & 23/06/2014, 15:41:25 & sun & 3356 \\
		Seq-4 & 23/06/2014, 15:36:04 & sun & 3438 \\
		Seq-5 & 23/06/2014, 15:14:44 & sun & 3690 \\
		Seq-6 & 24/06/2014, 14:15:17 & sun & 3065 \\
		Seq-7 & 24/06/2014, 14:09:07 & sun & 3285 \\
		Seq-8 & 24/06/2014, 14:20:41 & sun & 3678 \\
		\hline
		
	\end{tabular}
	\caption{Used sequences from the Oxford RobotCar dataset.}
	\label{tab:stat_robotcar}
\end{table*}

\subsection{Statistics of Oxford RobotCar dataset}
The statistics information of $8$ sequences we use in Oxford RobotCar \cite{maddern20171} is shown in Table \ref{tab:stat_robotcar}.

\begin{figure*}
	
	\centering
	\includegraphics[width=1\textwidth]{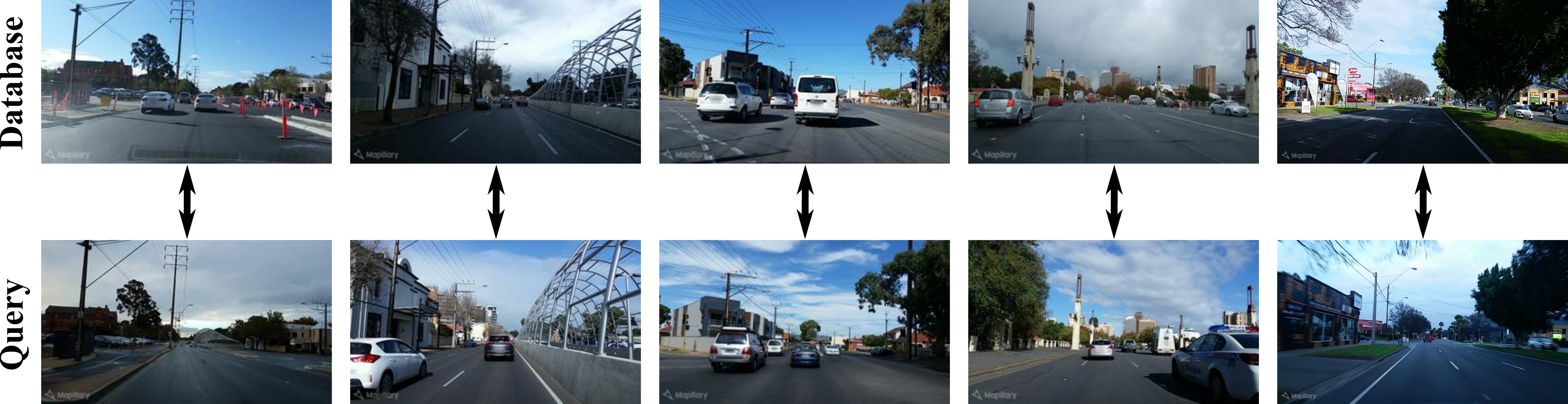}

	\caption{Samle images from our Mapillary dataset. The database image and its corresponding query have different appearance due to changes of environmental conditions and traffic density.}
	\label{fig:mapillary_example}
	\vspace{-0.2cm}
\end{figure*}

\subsection{Sample images of Mapillary}

Sample images from our Mapillary dataset are shown in Figure \ref{fig:mapillary_example}

\end{document}